\documentclass{article}

\usepackage{times}
\usepackage{color}

\usepackage{amsmath}
\usepackage{amsfonts}
\usepackage[pdftex]{graphicx}
\usepackage{url}

\usepackage{ijcai17}

\newcommand{\argmax}{\mathop{\rm arg~max}\limits}

\def\figures{ICF_AutoCapsule_disabled}
\def\evalfunc#1{Eval(#1)}

\title{Hierarchical Reinforcement Learning with Abductive Planning}

\author{
Kazeto Yamamoto$^{*\dag}$ \and Takashi Onishi$^{*\dag}$ \and Yoshimasa Tsuruoka$^{\dag\ddag}$ \\
\normalsize
$*$ Central Research Laboratories, NEC Corporation, Kanagawa, Japan \\
\small
E-mail: \{k-yamamoto@ft, t-onishi@bq\}.jp.nec.com \\
\normalsize
$\dag$ Artificial Intelligence Research Center,
National Institute of Advanced Industrial Science and Technology, Tokyo, Japan \\
\small
E-mail: \{kazeto.yamamoto, takashi.onishi, yoshimasa.tsuruoka\}@aist.go.jp \\
\normalsize
$\ddag$ The University of Tokyo \\
\small
E-mail: tsuruoka@logos.t.u-tokyo.ac.jp
}

\begin{document}
\maketitle

\begin{abstract}
 One of the key challenges in applying reinforcement learning to
 real-life problems is that the amount of train-and-error required to
 learn a good policy increases drastically as the task becomes complex.
 One potential solution to this problem is to combine reinforcement
 learning with automated symbol planning and utilize prior knowledge on
 the domain. However, existing methods have limitations in their
 applicability and expressiveness.
 In this paper we propose a hierarchical reinforcement learning method
 based on abductive symbolic planning.  The planner can deal with
 user-defined evaluation functions and is not based on the Herbrand
 theorem. Therefore it can utilize prior knowledge of the rewards and can
 work in a domain where the state space is unknown.
 We demonstrate empirically that our architecture significantly improves
 learning efficiency with respect to the amount of training examples on
 the evaluation domain, in which the state space is unknown and there
 exist multiple goals.
\end{abstract}


\section{Introduction}
\label{sec:introduction}

Reinforcement learning (RL) is a class of machine learning problems in
which an autonomous agent learns a policy to achieve a given task
through trial and error. Automated planning is an area of artificial
intelligence that studies how to make efficient plans for achieving a
given task with using predefined knowledge. Reinforcement learning and
automated planning are complementary to each other and various methods
to combine them have been
proposed~\cite{partalas08,grzes08,branavan12,konidaris14,konidaris15,leonetti16,andersen17}.

Partalas et al.~\shortcite{partalas08} classified those methods into two
categories: reinforcement learning to speed up automated planning and
reinforcement learning to increase domain knowledge for automated
planing.  In this paper, we focus on the former, and, more specifically,
address the problem of hierarchical reinforcement learning with
symbol-based planning. We are motivated by the fact that the lack of
efficiency in the learning process is an essential problem which has
hampered application of reinforcement learning to complicated domains in
the real world.

We argue that there are problems with the automated planners employed by
existing hierarchical reinforcement learning frameworks if they are to
be used in real-life applications.
First, most of the widely used planners and inference models are based on
the Herbrand theorem and thus need a set of constants as input to generate
a Herbrand universe. Therefore, the meaning representation used in planning
by such planners must satisfy the condition that, for each term of
the predicates, there exists a finite sequence of constant terms.  This
condition restricts the available meaning representations and often hampers
the application of symbolic planners to domains modeled by a partially
observable Markov decision process (POMDP).
Second, to the best of our knowledge, existing symbolic planners cannot
deal with pre-defined knowledge about rewards and hence cannot evaluate
the expectations of rewards in planning. Real-life problems often have
several goals with different priorities. For instance, the AI controller
of a plant may need to consider multiple goals in planning (e.g. safety
operation vs profit maximization).
Third, classical planners (e.g. STRIPS) can only deal with rules that
define actions.  Those planners cannot utilize other types of knowledge,
such as relationships of a subordinate concept to a main concept.
In order to address the above problems, we propose a new symbolic
planner that employs ILP-formulated abduction as its symbolic planner.
Abduction is a form of inference that is used to find the best
explanations to a given observation. The development of efficient
inference techniques for abduction in recent years warrants the
application of abduction with large knowledge bases to real-life
problems.  We show that our model can overcome the above issues through
experiments on a Minecraft-like evaluation task.

This paper consists of six sections.  In Section 2, we introduce the
formalism of reinforcement learning and automated reasoning in
abduction, and review some previous work on symbolic planning-based
hierarchical reinforcement learning. Section 3 describes our
abduction-based hierarchical reinforcement learning framework. Section 4
describes our evaluation domain. In Section 5, we report the results of
experiments.  The final section concludes the paper.


\section{Background}
\label{sec:background}

This section reviews related work on hierarchical reinforcement
learning with symbolic planning and abduction.

\subsection{Reinforcement Learning}

Reinforcement Learning is a subfield of machine learning that studies
how to build an autonomous agent that can learn a good behavior policy
through interactions with a given environment.  A problem of RL can be
formalized as a 4-tuple $\langle S, A, T, R \rangle$, where $S$ is a set
of propositional states, $A$ is a set of available actions, $T(s,a,s')
\rightarrow [0,1]$ is a function which defines the probability that
taking action $a \in A$ in state $s \in S$ will result in a transition
to state $s' \in S$, and $R(s,a,s') \rightarrow \mathbb{R}$ defines the
reward received when such a transition is made. The problem is called a
\textit{Markov Decision Process} (MDP) if the states are fully
observable; otherwise it is called a \textit{Partially Observable Markov
Decision Process} (POMDP).

The learning efficiency of RL decreases as the state space in the target
domain becomes larger.  This is a major problem in applying RL to large
real-life problems.  Although various approaches to solve this problem
have been proposed, we focus on methods that utilize symbolic automated
planners to improve the learning efficiency.

In automated planning, prior knowledge is used to produce plans which
would lead the world from its current state to its goal state.
Specially, Symbolic Automated Planning methods deal with symbolic rules
and generate symbolic plans.

Grounds \& Kudenko~\shortcite{grounds07} proposed \textit{PLANQ} to
improve the efficiency of RL in large-scale problems.  In PLANQ, a
STRIPS planner defines the abstract (high-level) behavior and a RL
component learns low-level behavior. PLANQ contains multiple Q-learning
agents for each high-level action. Each Q-learning agent learns the
behavior to achieve the abstract action corresponding to itself. The
authors have shown that a PLANQ-learner learns a good policy efficiently
through experiments on the domain where an agent moves in a grid world.

Grzes \& Kudenko~\shortcite{grzes08} proposed a reward shaping method,
in which a potential function considers high-level plans generated by a
STRIPS planner. In their method, a RL component (i.e. the low-level
planner) will receive intrinsic rewards when the agent follows the
high-level plan, where the amount of the intrinsic rewards is bigger as
the agent's state corresponds to the later state in the high-level
plan. They have shown that their approach helps the RL component to
learn a good policy efficiently through experiments on a navigation maze
problem.

\textit{DARLING}~\cite{leonetti16} is a model that can utilize a symbolic
automated planner to constrain the behavior of the agent to reasonable
choices. DARLING employs Answer Set Programming as an automated planner
in order to make their approach scalable enough to be applied to
real-life problems.

\subsection{Abduction}

Abduction (or abductive reasoning) is a form of inference to find the
best explanation to a given observation. More formally, logical
abduction is defined as follows:
\begin{description}
 \item[Given:] Background knowledge $B$, and observations $O$, where
	    $B$ and $O$ are sets of first-order logical formulas.
 \item[Find:] A \textit{hypothesis} (\textit{explanation}) $H$ such that
	    $H \cup B \models O$, $H \cup B \not\models \perp$, where
	    $H$ is a set of first-order logical formulas.
\end{description}
Typically, there exist several hypotheses $H$ that explain $O$. We call
each of them a \textit{candidate hypothesis}.  The goal of abduction is
to find the best hypothesis among candidate hypotheses according to a
specific evaluation measure.  Formally, we find $\hat{H} = \argmax_{H
\in \mathcal{H}} \evalfunc{H}$, where $\evalfunc{H}$ is a function
$\mathcal{H} \rightarrow \mathbb{R}$, which is called the
\textit{evaluation function}. The best hypothesis $\hat{H}$ is called
the \textit{solution hypothesis}. In the literature, several kinds of
evaluation functions have been
proposed~\cite{hobbs93,raghavan10,inoue12b,gordon16}.

Although abduction on first-order logic or similarly expressive formal
systems is computationally expensive, inference techniques developed in
recent years have improved its computational
efficiency~\cite{blythe11,inoue11,inoue12a,yamamoto15,inoue17}.
Specially, Inoue et al.~\shortcite{inoue11,inoue12a} proposed a method
(called \textit{ILP-formulated Abduction}) to formulate the process of
finding the solution hypothesis as a problem of Integer Linear
Programming (ILP) and showed that their method significantly improves
the computational efficiency of abduction.  In addition, since
ILP-formulated abduction is based on the directed acyclic graph
representation for abduction~\cite{charniak90} and thus generates a set
of candidate hypotheses in the manner similar to graph generation, it
does not need the grounding process.  This is a strong advantage in
comparison to other inference models based on the Herbrand theorem
(e.g. Answer Set Programming, Markov Logic
Networks~\cite{richardson06}).
More specifically, a set of candidate hypotheses is expressed as a
directed graph in which each node corresponds to a logical atom, where
each candidate hypothesis corresponds to a subset of nodes in the
directed graph. In the process to enumerate candidate hypotheses,
ILP-formulated abduction constructs the directed graph by applying two
kinds of operations to the observation: \textit{backward chaining} and
\textit{unification}.  Backward chaining is an operation that applies a
rule backward (i.e. consider that the presupposition may be true if the
consequence is true) and adds atoms in the presupposition to the graph.
Unification is an operation that unifies two atoms having the same
predicate and makes the assumption that each term of an atom is equal to
the corresponding term of the other atom. See Inoue and
Inui~\shortcite{inoue11} for details.

Abduction has been applied to various real-life problems such as
discourse
understanding~\cite{inoue12b,ovchinnikova13,sugiura13,gordon16},
question answering~\cite{and01,sasaki03} and automated
planning~\cite{shanahan00,pereira04}.  Many planning tasks can be
formulated as problems of abductive reasoning by giving an observation
consisting of the initial state and the goal state.  Abduction will find
the solution hypothesis explaining why the goal state has been achieved
by starting from the initial state.  Then the solution hypothesis can be
interpreted a plan from the initial state to the goal state.

\begin{figure}[htbp]
 \begin{center}
  \includegraphics[clip,width=8.0cm]{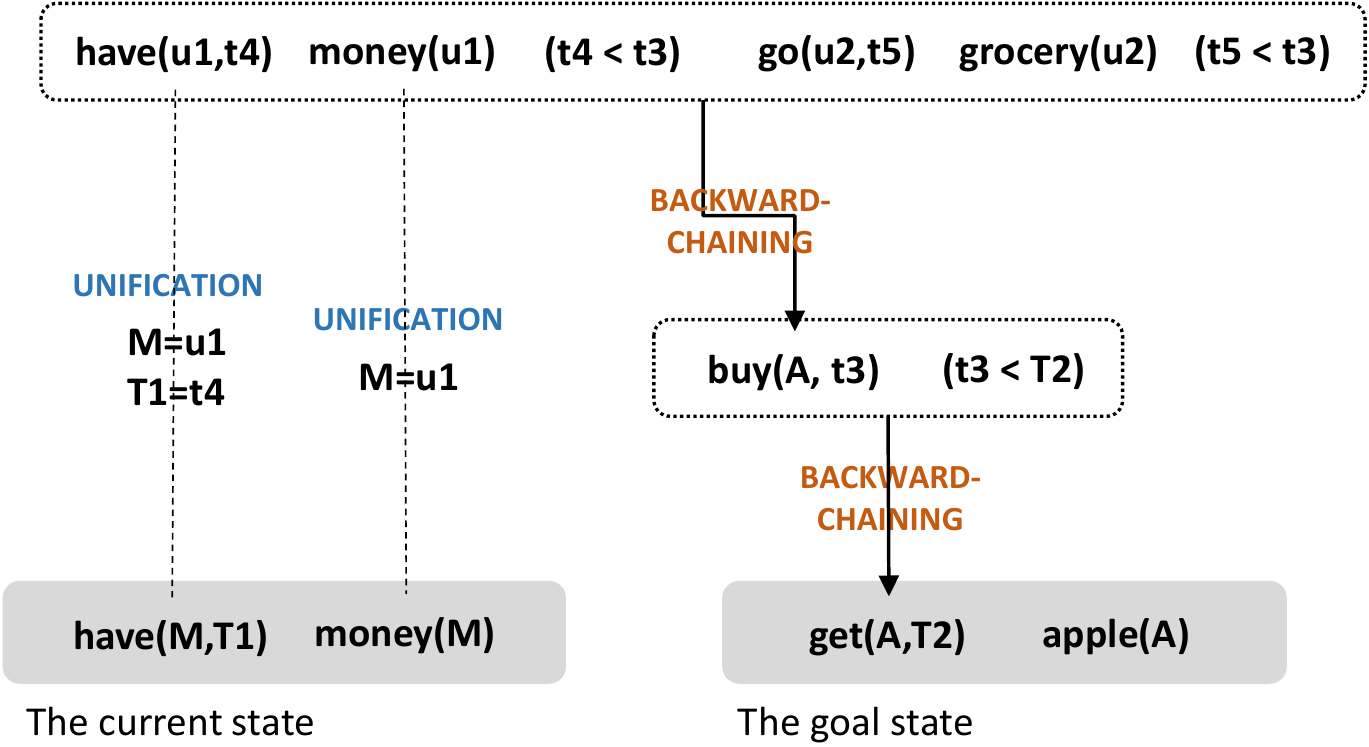}
  \caption{An example of a solution hypothesis generated by
  ILP-formulated abduction. Capitalized terms (e.g. $M$ and $T1$) are
  logical constants and the others (e.g. $u1$ and $t3$) are logical
  variables.  Atoms in a square are conjunctive (i.e. $have(M) \land
  money(M)$) and atoms in gray squares are observations. An equality
  between terms represents a relation between time points corresponding
  to the terms. Each solid, directed edge represents an operation of
  backward-chaining in which the tail atoms are hypothesized from the
  head atoms. Each dotted, undirected edge represents a
  unification. Each label on a unification edge such as $M=u1$ is an
  equality between arguments led by the unification.}
  \label{fig:plan}
 \end{center}
\end{figure}

Figure \ref{fig:plan} shows an example of a solution hypothesis by
ILP-formulated abduction in automated planning.  A solution hypothesis
is expressed as a directed acyclic graph and thus we can obtain richer
information about the inference than one of other inference
frameworks. From this graph, we can obtain the plan to get an apple,
namely go a grocery and buy an apple.


\section{Proposed Architecture}
\label{sec:model}

This section describes abduction-based hierarchical reinforcement learning.

\begin{figure}[htbp]
 \begin{center}
  \includegraphics[clip,width=8.0cm]{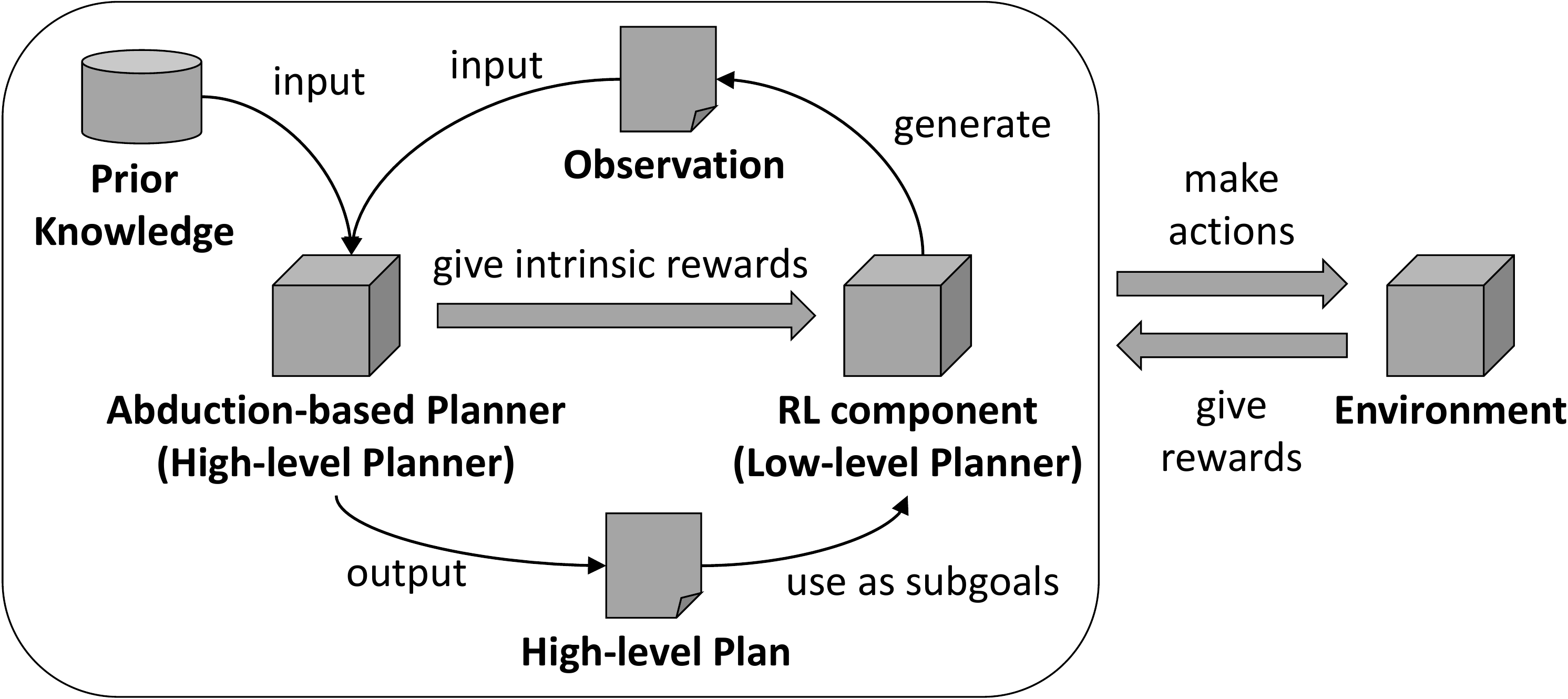}
  \caption{The basic structure of our architecture.}
  \label{fig:architecture}
 \end{center}
\end{figure}

Figure \ref{fig:architecture} shows the basic structure of our
architecture.  Using a predefined knowledge base, the abduction-based
symbolic planner generates plans at the abstract level.  The planner based
on a reinforcement learning model interprets the plans made by the
symbolic planner as a sequence of subgoals (options) and plans a
specific action on the next step.  This structure is similar to existing
hierarchical reinforcement learning methods based on symbolic planners,
such as PLANQ.  Following previous work, we call the abstraction level
on which the symbolic planner works \textbf{high-level} and the
abstraction level on which the planner based on the reinforcement
learning model works \textbf{low-level}. We use the term the {\it
high-level planner} to refer to the planner based on abduction, and the
term the {\it low-level planner} the planner based on the reinforcement
learning model.

Here we describe the algorithm for choosing an action.  First, the
system converts the current state and the goal state into an observation
in first-order logic for abduction.  Next, the system performs abduction
for this observation and then makes a high-level plan to achieve the
goal state.  We use a modified version of the evaluation function in
Weighted Abduction~\cite{hobbs93} in order to obtain a good plan.  We
describe the details of this evaluation function in Section
\ref{subsec:evalfunc}.  Finally, the system decides the next action by
considering the nearest subgoal in the high-level plan.  Following
hierarchical-DQN~\cite{kulkarni16}, the system gives intrinsic rewards
to the low-level component when the subgoal is completed, and thus the
low-level component will learn the behavior to achieve subgoals by
considering the intrinsic rewards.

One can use an arbitrary method to make a high-level plan from a solution
hypothesis in our architecture. In this paper, assuming that the graph
structure corresponds to the time order, we make a high-level plan from
actions sorted by distance from the goal state. For example, actions in
the solution hypothesis shown in Figure \ref{fig:plan} may be
\textsf{get-apple}, \textsf{buy-apple}, \textsf{have-money} and
\textsf{go-grocery}. Sorting them by distance from the goal state, we
can obtain a high-level plan \{ \textsf{go-grocery}, \textsf{buy-apple},
\textsf{get-apple} \}. The action \textsf{have-money} is excluded from
the high-level plan because it has been already satisfied by the current
state.

Using ILP-formulated abduction as the high level planner has several
benefits.
First, since ILP-formulated abduction does not need a set of constants
as input, our architecture can deal with a domain in which the size of
state space is unpredictable. In other words, there is no need to
consider whether the state space made from the current meaning
representation is a closed set or not. When using other logical
inference models, it is often hard to find which meaning representation
is appropriate for the target domain. This difficulty can be sidestepped
by ILP-formulated abduction.
Second, this advantage gives our architecture another benefit, namely
the ability to make plans of an arbitrary length. This is an advantage
over existing logical inference models (e.g. Answer Set Programming).
Third, an advantage over classical planners is the ability to use types
of knowledge other than action definitions. For instance, in STRIPS, one
cannot define rules of relations between objects (e.g. $coal(x)
\Rightarrow fuel(x)$).
Finally, ILP-formulated abduction provides directed graphs as the
solution hypothesis. Compared with other logical inference models which
just return sets of logical symbols as outputs, abduction can provide
more interpretable outputs.

\subsection{Evaluation Function}
\label{subsec:evalfunc}

In this section, we describe the evaluation function used in the
abductive planner of our architecture.

In general abduction, evaluation functions are used to evaluate the
plausibility of each hypothesis as the explanation for the observation.
For instance, the evaluation function of probabilistic abduction
(e.g. Etcetera Abduction~\cite{gordon16}) is the posterior probability
$P(H|O)$, where $H$ is a hypothesis and $O$ is the observation.

However, what we expect abduction to find in this work is not the most
probable one, but the most promising one as a high-level plan.  In other
words, our evaluation function needs to consider not only the possibility
of a hypothesis but also the reward that the agent will receive by
completing the plan made from the hypothesis.

Therefore, we add a new term of expected reward to the standard evaluation
function:
\begin{equation}
 \evalfunc{H} = E_{0}(H) + E_{R}(H),
\end{equation}
where $E_{0}(H)$ is some evaluation function in an existing abduction model,
such as Weighted Abduction and Etcetera Abduction.  $E_{R}(H)$ is the
task-specific function that evaluates the amount of reward on
completing a plan in hypothesis $H$.  More specifically, in this paper,
we use an evaluation function based on Weighted Abduction:
\begin{equation}
 \evalfunc{H} = -Cost(H) - r_H
\end{equation}
where $Cost(H)$ is the cost function in Weighted Abduction and $r_H$ is the
amount of reward of completing a plan in hypothesis $H$.

Although we employ Weighted Abduction as a base model due to the
availability of an efficient reasoning engine, a different abduction
model could be used.  For instance, using a probabilistic abduction
model, one can define an evaluation function to evaluate the exact
expectation of reward, namely $\evalfunc{H} = \log(P(H|O)) + \log(r_H)$.


\section{Evaluation Domain}
\label{sec:task}

This section describes the domain we used for evaluating our abduction-based RL method.

In this paper, we use a domain of grid-based virtual world based on
Minecraft. Each grid cell is either of land or lava and can contain
materials or utilities. The player can move around the world, pick up
materials and craft objects with utilities. Each episode will end when the
player arrives at the goal position, when the player walks into a lava-grid,
or when the player has executed 100 actions.

In order to examine the effectiveness of our approach empirically, we
set up the problem so that it has types of complexity that tend to exist
in real-world problems: partial observability, multiple goals, delayed
reward and multitask.

\begin{figure}[htbp]
 \begin{center}
  \includegraphics[clip,width=8.0cm]{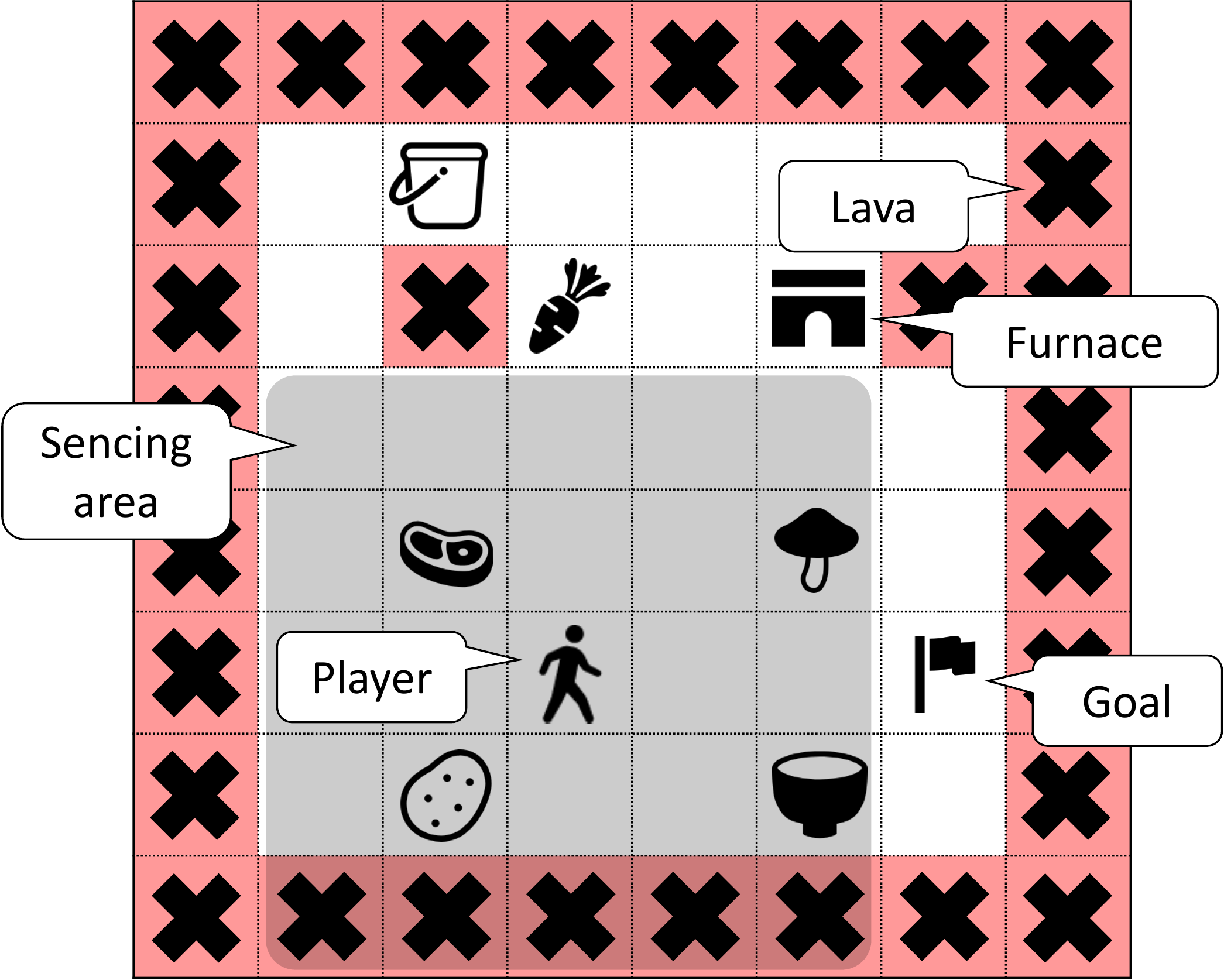}
  \caption{An example of tasks in our evaluation domain.}
  \label{fig:domain}
 \end{center}
\end{figure}

\begin{description}
 \item[Partial Observability]
	    The player at the initial state does not have any knowledge
	    of the environment. More specifically, he does not know the
	    size of the grid world, where he is, what items there are,
	    or the positions of materials and utilities in the grid
	    world. He can detect the existence of an object in the world
	    when he gets close to it. For example, the gray area around
	    the player in Figure \ref{fig:domain} shows the range of his
	    sensing.  Therefore, he knows nothing about the outside of
	    this area at the initial state.

 \item[Multiple Goals and Delayed Reward]
	    The player receives a reward only when he arrives at the
	    goal position. The amount of the reward depends on what he
	    can craft on arriving at the goal.  The reward will be high
	    if he can craft an object made from many materials.  For
	    instance, the reward given to a player who has enough
	    materials to cook \textsf{rabbit-stew} is much higher than
	    that for a player who has only collected \textsf{rabbit}.

 \item[Multitask]
	    In this domain, the layout of the grid world is randomized
	    on every episode.  Specifically, the player's starting
	    position, the goal position, the arrangement of lava, the
	    width of the grid-world, the variation of materials and
	    their positions vary randomly. The range of the width of the
	    grid-world $w$ is $12 \le w \le 15$. Each grid world
	    contains $4 \sim 9$ kinds of materials and is
	    always surrounded by lava.
\end{description}

It should be noted that, since the variation of materials in the world
may change, it is possible that the player cannot craft the optimal
object in some episodes.  In other words, the optimal goals of different
episodes are different.  Therefore, the player needs to judge which goal
is the most appropriate in each episode.
For example, the player will receive the highest reward when he can
cook \textsf{rabbit-stew}, which is made from \textsf{rabbit},
\textsf{bowl}, \textsf{mushroom}, \textsf{potato}, \textsf{carrot} and
some fuel to use a furnace.  Therefore, the player cannot cook it if any
of its materials does not exist in the world.

These characteristics make it difficult to apply existing reinforcement
learning models to this evaluation domain. The state space in this
domain is unpredictable and thus existing first-order symbolic
planners based on Herbrand's theorem, such as Answer Set Programming and
STRIPS, are not straightforwardly applicable to this domain.

Let us discuss in more detail the difficulty of applying Herbrand's
theorem-based planners to this domain. Since those planners need a set
of constants to make a Herbrand universe, one must define predicates so
that one can enumerate all possible arguments in advance. However, most
of the objects in this domain (e.g., grid cells, materials and time points) 
are not enumerable; that is, one cannot define closed sets of arguments
corresponding to those objects in advance.  Therefore one cannot avoid
giving a huge set of constants to deal with all possible cases or
abstracting predicates so that its argument set is known in advance.
The former may be computationally intractable and the latter may be too
time-consuming and difficult for human.

Following the conditions in General Game Playing~\cite{genesereth05}, we
had conducted evaluation on the following presuppositions.  First, the
player can use prior knowledge of the dynamics of the target
domain. That includes knowledge of crafting rules and the amount of
reward for each object.  Second, the player cannot use the knowledge of
task-specific strategies for the target domain. In other words, we do
not add any knowledge of how to move for getting higher rewards.


\section{Experiments}
\label{sec:experiments}

We evaluated our approach in the domain described in Section \ref{sec:task}.

We compared the following three models. First, \textbf{NO-PLANNER} is a
RL model without a high-level planner. Second, \textbf{FIXED-GOAL} is
the model in which the high-level planner always makes plans so that the
player achieve the most ideal goal (i.e. cooking
\textsf{rabbit-stew}). We consider this model to correspond to existing
symbolic planner-based hierarchical RL models, in which the high-level
planners cannot deal with prior knowledge of the rewards.  Finally,
\textbf{ABDUCTIVE} is our proposed model, in which the high-level
planner is based on abduction.

We employed Proximal Policy Optimization algorithm
(PPO)~\cite{schulman17}\footnote{We used the library implemented by
OpenAI, available at \url{https://github.com/openai/baselines}} as the
low-level component for each model. PPO is a state-of-the-art policy
gradient RL algorithm, and is relatively easy to implement.

In order to perform abduction based on our evaluation function proposed
in Section \ref{subsec:evalfunc}, we implemented a modified version of
Phillip\footnote{It is available at
\url{https://github.com/kazeto/phillip}}, a state-of-the-art engine for
ILP-formulated abductive reasoning, and used it for the high-level
planner of ABDUCTIVE. In order to improve the time efficiency of
planning, we cached the inference results for each observation and
reused them whenever possible.

\begin{figure}[htbp]
 \begin{center}
  \includegraphics[clip,width=8.0cm]{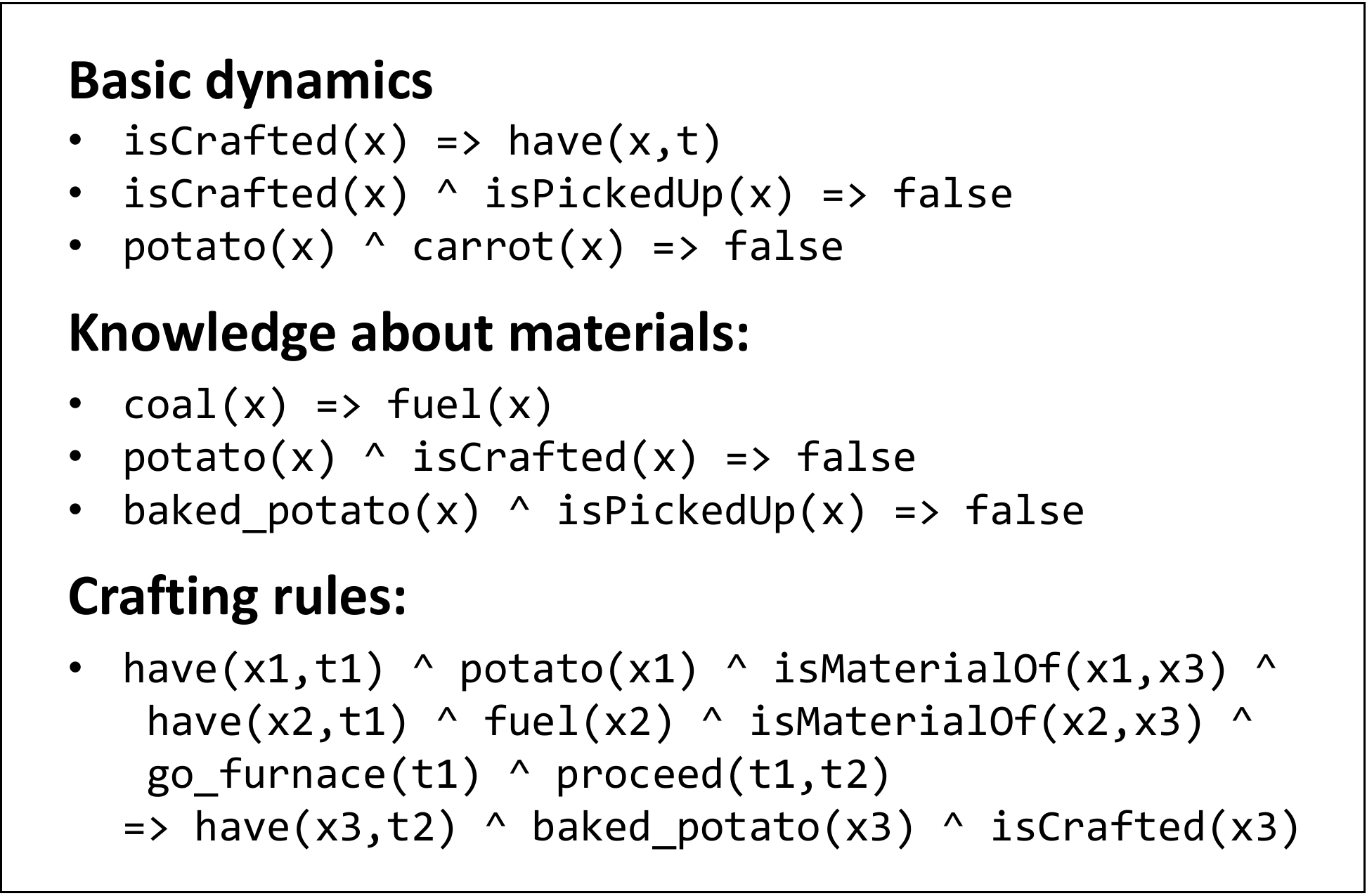}
  \caption{Examples of rules that we used.}
  \label{fig:rules}
 \end{center}
\end{figure}

We manually constructed the prior knowledge of the evaluation domain for the
high-level planner. The knowledge base contains 31 predicates and 125
rules.  As stated in Section \ref{sec:task}, these rules consist of only
the ones for the dynamics of the domain, such as crafting rules and
properties of objects.  We describe examples of the rules in Figure
\ref{fig:rules}.

We used roughly three types of actions as elements in high-level plans,
namely finding a certain object, picking up a certain material and going
to a certain place. Each of them takes one argument
(e.g. \textsf{get-rabbit}) and thus we actually used 20 actions in a
high-level plan. For instance, a high-level planner in this experiment
may generate high-level plans like \{ \textsf{find-coal},
\textsf{get-coal}, \textsf{get-rabbit}, \textsf{go-furnace},
\textsf{go-goal} \}.

As stated previously, the content of each task is generated randomly.
Since we use the number of episodes as the random number seed for task
generation, the set of tasks in each experiment is the same.  We tested
the model performance every 10 episodes in each experiment.

\subsection*{Experimental Result}

\begin{figure}[htbp]
 \begin{center}
  \includegraphics[clip,width=8.0cm]{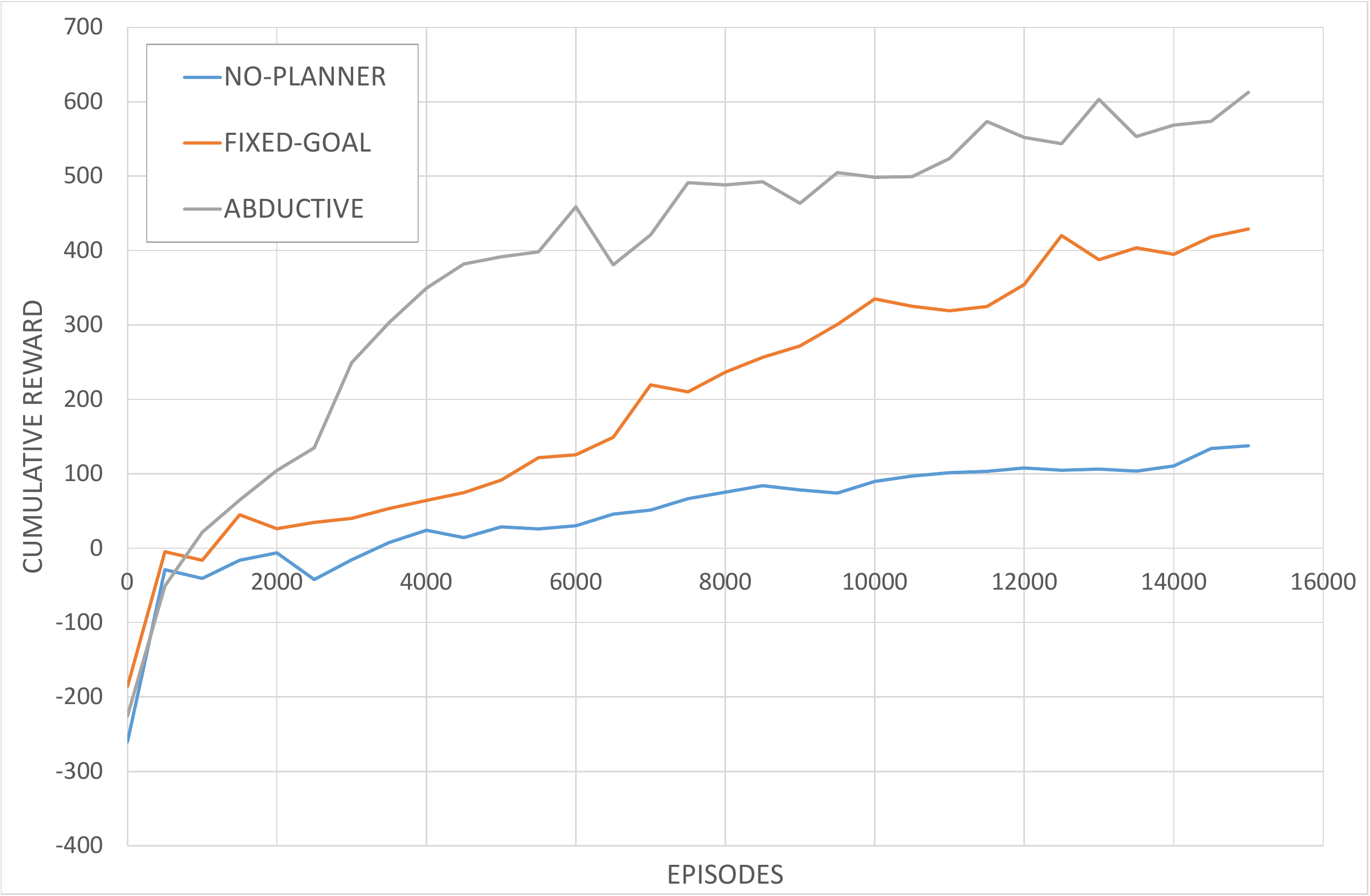}
  \caption{The performance of three models.}
  \label{fig:result}
 \end{center}
\end{figure}

Figure \ref{fig:result} illustrates the curves of the cumulative rewards
averaged over 5 independent trials for each model.  Each plot is
averaged over a sliding window of 500 episodes.

As we can see, ABDUCTIVE obtained much more rewards than other models
and learned more efficiently with respect to the number of training
examples.  Learning efficiency is important when RL is applied to
real-life problems.  In such domains, the time required for trial and
error can be prohibitively long because of the computational cost of a
simulator or necessity of manual operations.

\begin{figure}[htbp]
 \begin{center}
  \includegraphics[clip,width=8.0cm]{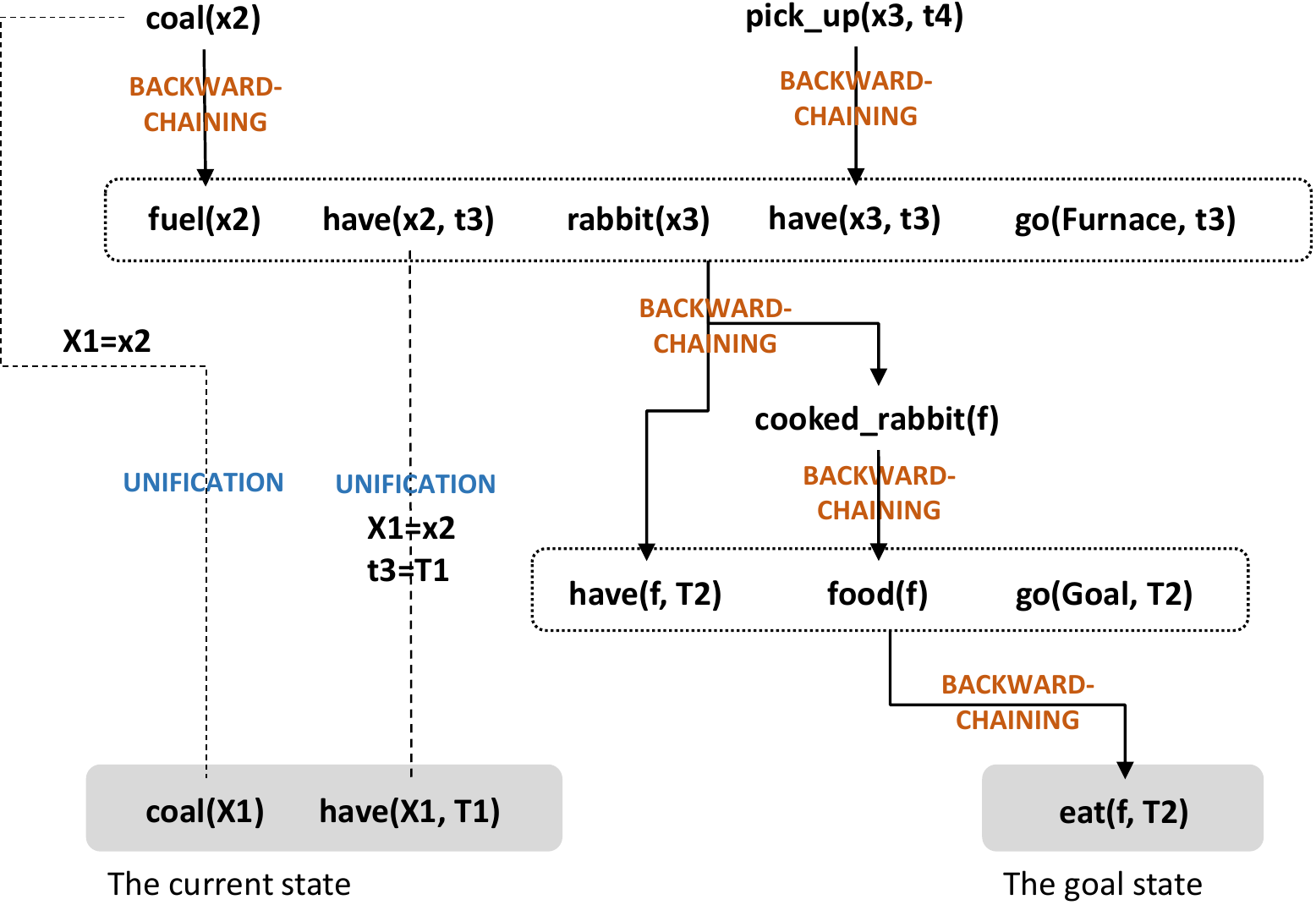}
  \caption{An example of solution hypotheses generated by the abductive
  planner in the evaluation domain. }
  \label{fig:proof}
 \end{center}
\end{figure}

Figure \ref{fig:proof} is an example of solution hypotheses made by the
abductive planner in the evaluation domain. Our architecture may convert
this into a subgoal sequence --- pick up \textsf{rabbit}, go to
\textsf{furnace} and go to \textsf{goal}.  As we can see, since it is
described as a graph how the planner inferred the plan, our system can
improve interpretability of the content of the inference. From this
proof graph, we can see that our planner can make plans of an arbitrary
length and can use types of knowledge other than action definitions.

One limitation of our architecture is the large variance in CPU time
required per time step. Most of the action selections can be made in a
few milliseconds, but when the high-level planner needs to perform
abductive planning, the selection may take a few seconds.  For this
issue, there are several directions of future work to reduce the
frequency of performing abduction. One is to improve the algorithm for
finding reusable cached results. Our current implementation uses cached
results only when exactly the same observation is given. The
computational cost of high-level planning could be significantly reduced
if the high-level planner can reuse a cache for similar observations as
well.


\section{Conclusion}
\label{sec:conclusion}

We proposed an architecture of abduction-based hierarchical
reinforcement learning and demonstrated that it improves the efficiency of
reinforcement learning in a complex domain.

Our ILP-formulated abduction-based symbolic planner is not based on the Herbrand
theorem and thus can work in domains where the state space is unknown.
Moreover, since it can deal with various evaluation functions including
user-defined ones, we can easily allow an abductive planner to utilize
prior knowledge about rewards.

In future work, we plan to employ machine learning methods for
abduction.  In recent years, some methods for machine learning of
abduction have been proposed~\cite{yamamoto13,inoue12b}.  Although we
manually made prior knowledge for the experiments in this work, we could
apply these methods to our architecture.  Specifically, if we could
divide the errors into the high-level component's errors and the
low-level component's errors, we can update the weights of symbolic
rules used in the abductive planner discriminatively when the high-level
planner fails.



\bibliographystyle{named}
\bibliography{main}

\end{document}